# Operationalization of Scenario-Based Safety Assessment of Automated Driving Systems

Olaf Op den Camp
*Integrated Vehicle Safety*
TNO
Helmond, the Netherlands
0000-0002-6355-134X

Erwin de Gelder
*Integrated Vehicle Safety*
TNO
Helmond, the Netherlands
0000-0003-4260-4294

*Abstract*— Before introducing an Automated Driving System (ADS) on the road at scale, the manufacturer must conduct some sort of safety assurance. To structure and harmonize the safety assurance process, the UNECE WP.29 Working Party on Automated/Autonomous and Connected Vehicles (GRVA) is developing the New Assessment/Test Method (NATM) that indicates <u>what</u> steps need to be taken for safety assessment of an ADS. In this paper, we will show <u>how</u> to practically conduct safety assessment making use of a scenario database, and what additional steps must be taken to fully operationalize the NATM. In addition, we will elaborate on how the use of scenario databases fits with methods developed in the Horizon Europe projects that focus on safety assessment following the NATM approach.

## I. Introduction

A safety assurance process that is conducted by the manufacturer before introducing an Automated Driving System (ADS), intends to assure that the ADS responds appropriately in all situations it is designed for and that the ADS is able to avoid any reasonably foreseeable and reasonably preventable collisions. The information out of the safety assurance process is not only important for manufacturers, but also for authorities that have the responsibility to guard the safety of their citizens in traffic. Safety assurance is most important for consumers (and fleet owners) using an ADS with the expectation that the system is safe, reliable, and trustworthy.

To structure and harmonize this process, the UNECE WP.29 Working Party on Automated/Autonomous and Connected Vehicles (GRVA) is developing the New Assessment/Test Method (NATM) [1], which is already recognized across many countries (e.g., Japan, South Korea, the EU and the USA). The NATM considers a multi-pillar, scenario-based approach, where tests (virtual, physical, and real-world) are based on scenarios considering the ADS's Operational Design Domain (ODD) and requirements (Fig. 1). The real-world scenarios that feed into the safety assessment framework are knowledge based and/or data driven, ranging from accident scenarios to everyday driving scenarios. The GRVA does not, however, provide methods or tools to operationalize the NATM framework. Thus, to allow stakeholders to perform safety assessment following the NATM approach, methods and tools are needed to operationalize it.

The research presented in this paper has been made possible by the Horizon Europe project SYNERGIES. This project has been funded by the European Union under grant agreement No. 101146542. Views and opinions expressed are however those of the author(s) only and do not necessarily reflect those of the European Union or the European Climate, Infrastructure and Environment Executive Agency (CINEA). Neither the European Union nor the granting authority can be held responsible for them.

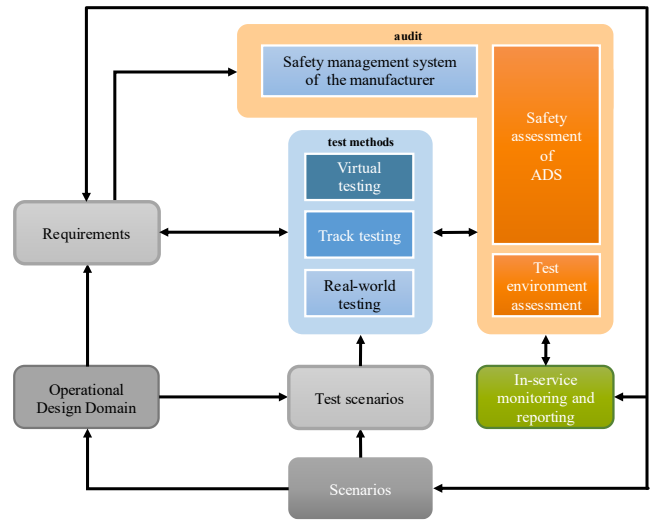

Fig. 1. The NATM multi-pillar approach. Figure is adapted from [2] by making a distinction between (real-world) scenarios and test scenarios.

In this paper, we will show how to use a scenario database for safety assessment, and what additional steps must be taken to fully operationalize the NATM and to come to an efficient and acceptable safety assessment process. In Section II, an introduction to the safety assessment framework according to NATM is provided, indicating the different steps that are foreseen to be needed to assure safe introduction of an ADS. One of the largest challenges in the practical application is the selection of a set of scenarios that sufficiently covers the ADS's ODD. In Section III, a federated approach is proposed to deal with this challenge. Section IV provides a view on additional aspects that need to be considered for the practical application of NATM. In Section V, conclusions are drawn and recommendations are provided for further research on this topic.

## II. Safety Assessment Framework

### A. Multi-pillar approach

The NATM applies a multi-pillar approach, with three testing pillars: virtual testing using computer models and simulations, physical testing on a proving ground, and real-world testing on the public road. The fourth pillar considers an auditing process of the followed procedures, the testing tools/methods, and the Safety Management System of the ADS manufacturer. In-service monitoring and deployment when the ADS is deployed on the public road is considered the fifth and final pillar. In European projects such as HEADSTART [3], SUNRISE [4] and SYNERGIES [5], the focus has mainly been on the three testing pillars.

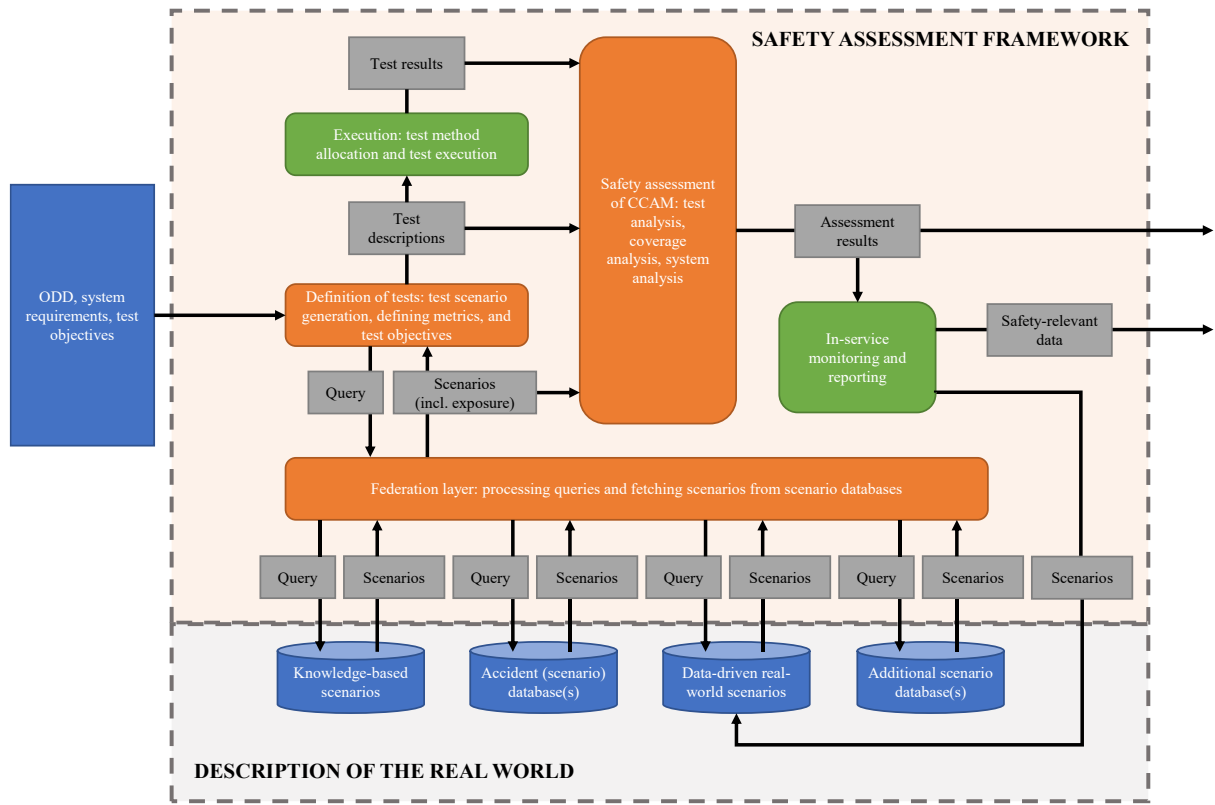

Fig. 2. Operationalized scenario-based safety assessment framework that is in agreement with the NATM multi-pillar approach. Different types of scenario databases are used to determine the required test scenarios as input for testing.

NATM takes scenarios as a starting point. *Scenarios* describe any situation on the road that an ADS might encounter in operation during its lifetime, form the basis to describe the ODD of the ADS. Scenarios, which are agnostic of the ADS technology, are typically stored in scenario databases. Examples of scenario databases are SAFETYPOOL [6], STREETWISE [7], ADSCENE [8], and SCENARIO.CENTER [9]. Scenarios can be used to describe a system's ODD, whereas tests need to provide evidence that the ADS responds to the scenarios within its ODD appropriately and according to requirements.

*Test scenarios* form the input to the different testing methods. It is common to determine test scenarios based on the scenarios that are relevant to the ADS under test and a specification of the ADS. Often challenging scenarios for the ADS under test are selected, but one should also evaluate the response of the ADS for scenarios that are rather common to get an understanding of the performance of the system across its ODD.

*B. The NATM framework in practice*

We have drafted a safety assessment framework (SAF, Fig. 2) to operationalize the multi-pillar approach as proposed by UNECE [7]. The SAF includes, but is not limited to, processes to:

- *Generate the relevant test scenarios* Based on a description of the ODD and the system requirements (incl. requirements resulting from a description of the dynamic driving task), the set of relevant test scenarios is determined.

- *Search scenario databases using a federation layer* Scenario databases may have different origins, based on the way information is collected in the database. We distinguish knowledge-based scenarios and data-driven scenarios. Data-driven scenarios may result from accident databases and from data collected during everyday driving. In the Horizon Europe project SUNRISE [4], a federation approach has been developed to make it possible to access different heterogeneous scenario databases from a single point of access. In SYNERGIES [5], this federated approach will be further extended for traceability of the scenario selection, adding quality metrics to the scenario set with regards to completeness of the data and coverage of the ODD.

- *Allocate test scenarios to the different test methods and execute tests* Virtual testing is very promising, as it is capable to evaluate system performance in a large number of test scenarios rapidly and safely. Validation of the used models in virtual simulation is a challenge though, and physical testing on a test track or on the road are required to validate and complement the virtual tests.

- *Analyse and assess the test results* The results of the tests need to be evaluated to draw a conclusion regarding the safety of the ADS. A check needs to be made whether the system provides an appropriate response in all test scenarios and whether the test scenarios cover the ODD sufficiently. Based on the analysis result, or the evidence regarding the safety of the system provided by the manufacturer and subsequent check by the authorities, the authority may approve the deployment of the system onto the public road, e.g., for large-scale testing.

- The NATM framework also anticipates the monitoring of ADS that are deployed on the road. This is to receive early feedback of system performance during large scale testing or deployment. In the EU project

CERTAIN, methods for *In-service monitoring and reporting* (ISMR) will be developed.

### III. SCENARIO DATABASES IN A FEDERATED APPROACH

The number of possible scenarios that an ADS in operation may encounter on the road during its lifetime is virtually infinite. Additionally, there are many variations to these scenarios resulting from the variability of environmental conditions, weather influences and lighting conditions, which may be country specific. There are several key challenges in the collection of scenarios. First, it is the identification of scenarios that are representative for the ODD of the ADS. Second, we need to make sure that we collected enough scenarios to cover the entire ODD of the system – this includes those scenarios with a low probability of occurrence. Only by collaborative efforts, it is possible to address these challenges.

In current practices, individual organizations or consortia collect scenarios from a limited area (e.g., a region or a country) in individual scenario databases, each having their own structure and scenario formats, often dedicated to a specific application. The SUNRISE project has developed a federated approach, that facilitates the collaboration of scenario database hosts. Through a federation layer, target users are able to access the different scenario databases to generate a set of relevant test scenarios as input to safety assessment. The federation layer enables the collaboration with different types of heterogeneous scenario databases using a common interface between the scenario database and the federation layer. This is one solution for increasing the scenario coverage of an ODD in a collaborative effort.

The SYNERGIES project takes the SUNRISE federation layer and develops it further to achieve an increased coverage and representativeness of scenarios for the generation of test scenarios. It considers the fact that individual scenario databases are continuously updated, from different sources, with different data types, and with different applications in mind. To support the governance of the federated approach and enhance collaboration between scenario databases, SYNERGIES develops a methodology and tools to determine metrics for scenario representativeness and coverage. This allows target users to get an understanding of the 'quality' of a selected set of test scenarios for a given ODD that results from a scenario selection involving multiple scenario databases. In this way, users can use the network of collaborating scenario databases as if it is one EU-wide scenario dataspace. Depending on the ODD for which scenario-based tests are needed, a search through the scenario dataspace can be issued. Additionally, SYNERGIES develops tools to document the search and the results, including federation-layer provided outcomes, with the necessary metadata to reconstruct the search for later reference.

A federated approach additionally allows the easy extension of the EU scenario dataspace with new scenario database initiatives. Only an interface has to be made between the added scenario database and the federation layer. Recommendations for developing an additional scenario database and guidelines for making the interface with the federation layer have been provided by the SUNRISE project.

### IV. ADDITIONAL ASPECTS FOR PRACTICAL APPLICATION

The objective of the SAF (following the NATM approach) is to come to a safety case for the ADS under consideration, for the manufacturer to take a decision whether the ADS is ready for an assessment by the type-approval authority. The safety case is provided by the manufacturer to the type-approval authority, for them to decide whether the ADS is meeting the legal safety requirements and can be allowed for deployment onto the public road.

Fast changing legislation and the large innovation speed of ADS requires a feasible yet thorough safety assessment process. In the next sections, considerations and first principles are provided on the use of the SAF.

#### A. Acceptable Means of Compliance (AMC)

AMCs are non-binding guidelines or standard procedures that are issued by regulatory bodies or (vehicle) authorities to help manufacturers to demonstrate compliance with the regulations [10]. Following AMCs is not mandatory. However, it provides a recognized and accepted way to meet regulatory requirements. Since AMCs are non-binding, manufacturers may choose alternative means to comply with the regulations. In that case, the manufacturer should demonstrate an equivalent level of safety or compliance with the regulations. Regulatory bodies or authorities then assess these alternative approaches on a case-by-case basis. In case of agreement and approval of the alternative approach, additional AMCs may be formulated by the authorities or regulators.

We mention the concept of AMCs as this is the current practice for safety assurance in relation to regulations and the approval of ADS. The reason is that regulations are not carved in stone and change on a regular basis, and hence also the ways in which compliance with these regulations needs to be shown is subject to change. The SAF provides a useful process for setting up a safety case, and consequently it supports the definition of AMCs.

#### B. Regulations and the Safety Case

Regulations typically consist of different type of requirements (examples are taken from UNECE R157 [11] [12]):

- *Specific requirements* (e.g., requirements on the specific allowed following distance, Clause 5.2.3.3.) that need to be checked in a large number of possible occurring scenarios. Often, the metric and a reference to judge compliance with a specific requirement is provided in the regulations as well. This makes it possible to determine whether the ADS shows a pass or fail in each test case that is set up for checking the specific requirement.

- *General (but not less important) requirements*, e.g., the ADS shall not cause any collisions that are reasonably foreseeable and preventable (Clause 5.1.1.). To evaluate the general requirement of the given example, it needs to be shown what collisions may be considered reasonably foreseeable and reasonably preventable, and that the ADS under consideration is indeed able to avoid a collision in any of the relevant scenarios.

- *Soft requirements*, e.g., the ADS shall aim to keep the vehicle in a stable lateral and longitudinal motion inside the lane of travel to avoid confusing other road users (Clause 5.2.1.). Though soft requirements need to be fulfilled as well, no strict metrics and references are provided, and the regulation is open how to evaluate such requirements.

In a safety case, information is collected to substantiate the hypothesis that the ADS meets the legal safety requirements and hence can be deployed safely. Such information consists of (but is not limited to):

- *Test matrix per requirement* How well specific requirements are met can be shown by providing the complete test matrix of tests performed per requirement. For each test case in the matrix, the result of the test is indicated, and by comparing the test result with the appropriate reference, it is indicated whether the ADS passed or failed the test. In addition to providing the test matrix, it should also be shown how the test matrix has been compiled. Important information in that respect considers the coverage of the set of test scenarios with respect to the ODD of the ADS.

- *Convincing argumentation* Dealing with general requirements in the safety case not only requires the provision of relevant information, but also of convincing argumentation that is supported with quantified evidence. For determining reasonably foreseeable collisions (from the example given under *general requirement*), the scenarios that fall within this category have to be identified. Ensuring that an ADS can safely deal with those scenarios, it is assured that the ADS avoids reasonably foreseeable collisions. In [13], it is shown how statistics are used to determine what scenarios can be considered reasonably foreseeable given the ODD of an ADS. Additionally, it is shown how to study the extent to which a collision is preventable for each individual scenario. This approach involves comparing how an ADS can safely handle each scenario that a human driver can manage safely. The quantification of what collisions can be prevented by a safe and competent human driver, is an important research topic.

- *Description of response to scenarios relevant for the requirements* For 'soft' requirements, the manufacturer is asked to showcase how the system responds to the scenarios relevant for these requirements. The manufacturer should indicate how it has been evaluated (e.g. with tests, possibly in simulation) that the system behaves according to the soft requirements. Evidence can be provided via e.g. test reports. This allows authorities to check such evidence, e.g. by evaluating the test reports and performing spot checks [14].

### C. Quantification of Safety Risk using Statistics

Road authorities express a clear need of establishing a framework for the type approval and safety assessment that is fair, explainable, and understandable. Though vehicle systems are complex and the assessment procedure might be complicated, the assessment results should be unambiguous, easily understood by experts in the field, and explainable to politicians and the general public. An important metric in such a framework is the residual safety risk when a vehicle is allowed onto the road. The concept of risk is widely understood, and basing the safety assessment on that concept helps to come to a fair and acceptable assessment process.

There are several ways of using a risk estimate for an ADS, such as, e.g., Positive Risk Balance (PRB) and Globalement Au Moins Aussi Bon (GAMAB). The PRB states that the risk of an ADS-equipped vehicle should be lower than a state of the art human-driven vehicle in terms of injuries and fatalities [15]. GAMAB (translated as Globally As Least As Good) requires a new system to exceed the safety and performance levels of existing or previous systems [16].

Scenarios, having proper scenario statistics in mind during the collection process, form a good basis for the quantification of safety risk. When scenarios are identified from a continuous dataflow uploaded from vehicles that drive many kilometres on the road, it is possible to determine exposure values for such scenario and the scenario parameters using scenario statistics. These scenarios subsequently can be used to generate test scenarios as input to virtual tests, e.g. computer simulations. Such simulations can be used to determine the crash probabilities and the consequences of possible crashes for vehicle occupants and other road users being part of the crash. By integrating the product of crash probability, the consequences of the crash (e.g., as an injury severity) and the exposure of the scenario over all scenarios within the ODD, the safety risk induced by the ADS can be estimated. More information on procedures to estimate safety risk and a confidence interval for this estimate is found in [17].

### D. Validated Test Environments

Results of tests that contribute to evidence in a safety case only have value when it comes with a validation of the methods and tools that are used. Documentation of the followed methods, and references to the used tools, models, and data must be provided [18]. This includes documentation on results of the validation process of the tools and models. At least, the following methods and tools must be addressed:

- Scenario database and tools for selecting scenarios;

- Tools for the generation of testing scenarios (or conversion of scenarios into test scenarios), methods for the estimation of scenario parameter distributions, and sampling methods;

- Tools used in proving ground testing, such as vehicles, targets, view-blocking obstructions, localisation and measurement equipment, etc.;

- Simulation tools, simulation structure, and models used in virtual simulation such as sensor models, vehicle models, human models, test environment and condition models [19].

If it is not possible to refer to documentation on the validation of certain methods/tools, the validation of those methods/tools needs to be performed to complement the safety case with the results of the validation. How to conduct validation processes is considered outside the scope of this paper.

## V. DISCUSSION, CONCLUSION & RECOMMENDATIONS

In this paper, we have discussed aspects to consider when planning and conducting a safety assessment of automated driving systems, following the UNECE NATM multi-pillar approach. The considerations provided in this paper are however not complete.

In this paper, we have focused on generating a set of tests that sufficiently covers the ODD of the ADS that is being assessed. And though the need for good references (or acceptance criteria) was acknowledged for setting up a trustworthy safety case, it was not discussed how to establish such references. For appropriate safety assessment we do need clear references and acceptance criteria in addition to the tests that are proposed following the SAF [20]. Again, regulations are helpful. In the Commission Implementing Regulation on Type Approval of ADS [21], it is mentioned that the ADS should be free from 'unreasonable risk' and the ADS-driven vehicle is compared to a 'competent and carefully driven manual vehicle'.

The question is how to convert these qualitative references into quantitative acceptance criteria. Tejada et al. present a practical methodology for defining and quantifying competent driving [22]. It is based on mathematical formulas, called assertions, that capture the relationship among traffic participants and/or elements of driving behaviour. A driver is considered competent when maintaining the applicable relationships within 'acceptable' ranges at any given time. A type of driving study is introduced in which expert driver assessors record their opinion about the quality of driving of a human driver during driving sessions. In [22], a proof of concept is provided for quantifying acceptable behaviour for keeping distance in highway car-following scenarios. To provide quantitative acceptance criteria for the large variety of scenarios, considering what is accepted as competent driving for the different countries and driving cultures across Europe, requires a significant scaling up of this methodology.

Another aspect that was not considered in this paper concerns the safety performance once the ADS is deployed on the road after certification or type approval. The In-Service Monitoring and Reporting (ISMR) component mentioned in the NATM approach (see Fig. 1) enables a continuous safety validation that monitors and reports the system's safety state over the complete lifecycle, also after the system is deployed into operation and possibly is being upgraded by an over-the-air (OTA) software update. The Horizon Europe project CERTAIN (2025 – 2028) will develop an operationalized continuous safety assessment methodology. The ISMR methodology should be able to:

- Detect unexpected CCAM responses in daily operation. The methodology should provide an 'early' indication in case the system response deviates from the response that is expected from the system that has successfully passed the pre-deployment safety assessment for type approval.
- Detect 'unknown' scenarios in daily operation. In pre-deployment safety assurance, the manufacturer has to show that the ODD is sufficiently covered by representative scenarios. However, the actual operational domain may slightly differ from the ODD. Moreover, as the traffic system continues to change, e.g., with the introduction of new mobility devices and changes in the interaction with other road users, new previously unknown scenarios may occur.

With the introduction of ISMR at a large scale, the safety assurance of ADSs will be extended to the whole lifecycle of such systems, following all pillars of the NATM approach.

As the NATM shows, a proper safety assessment starts with achieving a good view on the large variety of scenarios that an ADS in operation possibly encounters on the road. The ADS needs to respond appropriately in each of these scenarios. In various EU projects, methods and tools for the identification, characterisation, collection and sharing of scenarios have been developed to make cooperation and collaboration between authorities, manufacturers, technology providers, and research organisations technically possible keeping intellectual property rights, data confidentiality and privacy regulations in mind. A major step is made in the SYNERGIES project, in establishing a scenario dataspace connecting various heterogeneous scenario databases across Europe and providing quality metrics for the evaluation of scenarios selected from the connected scenario databases as if it is one European scenario database. Proper governance of the developed tools and methods needs to be arranged beyond the SYNERGIES project, which will be concluded in 2027, to enable continuous collaboration between current scenario database hosts and to allow new scenario database initiatives to easily connect to the established network. Only through international cooperation and intensive collaboration between all stakeholders, we will be able to come to a feasible and widely accepted process for safety assessment that can be used at scale to accelerate the safe and responsible deployment of ADSs on public roads.